\documentclass[10pt,twocolumn,letterpaper]{article}

\usepackage{iccv}
\usepackage{times}
\usepackage{epsfig}
\usepackage{graphicx}
\usepackage{amsmath}
\usepackage{amssymb}

\usepackage{algorithm}
\usepackage{algorithmicx}
\usepackage[noend]{algpseudocode}
\usepackage{multirow}
\usepackage{float}
\usepackage[breaklinks=true,bookmarks=false]{hyperref}

\iccvfinalcopy 

\def\iccvPaperID{****} 

\ificcvfinal\fi

\begin{document}

\title{Unlimited-Size Diffusion Restoration}

\author{Yinhuai~Wang\textsuperscript{\rm 1}\quad
Jiwen~Yu\textsuperscript{\rm 1}\quad
Runyi~Yu\textsuperscript{\rm 1}\quad
Jian~Zhang\textsuperscript{\rm 1,2}\\
\small{\textsuperscript{\rm 1}Peking University, SECE\quad \textsuperscript{\rm 2}Peng Cheng Laboratory}\\
}

\ificcvfinal\thispagestyle{empty}\fi

\maketitle

\begin{abstract}
Recently, using diffusion models for zero-shot image restoration (IR) has become a new hot paradigm. This type of method only needs to use the pre-trained off-the-shelf diffusion models, without any finetuning, and can directly handle various IR tasks. The upper limit of the restoration performance depends on the pre-trained diffusion models, which are in rapid evolution. However, current methods only discuss how to deal with fixed-size images, but dealing with images of arbitrary sizes is very important for practical applications. This paper focuses on how to use those diffusion-based zero-shot IR methods to deal with any size while maintaining the excellent characteristics of zero-shot. A simple way to solve arbitrary size is to divide it into fixed-size patches and solve each patch independently. But this may yield significant artifacts since it neither considers the global semantics of all patches nor the local information of adjacent patches. Inspired by the Range-Null space Decomposition, we propose the Mask-Shift Restoration to address local incoherence and propose the Hierarchical Restoration to alleviate out-of-domain issues. Our simple, parameter-free approaches can be used not only for image restoration but also for image generation of unlimited sizes, with the potential to be a general tool for diffusion models. Code: \textcolor{blue}{https://github.com/wyhuai/DDNM/tree/main/hq\_demo}.
\end{abstract}

\begin{figure}[t]
  \centering
  \includegraphics[width=1\linewidth]{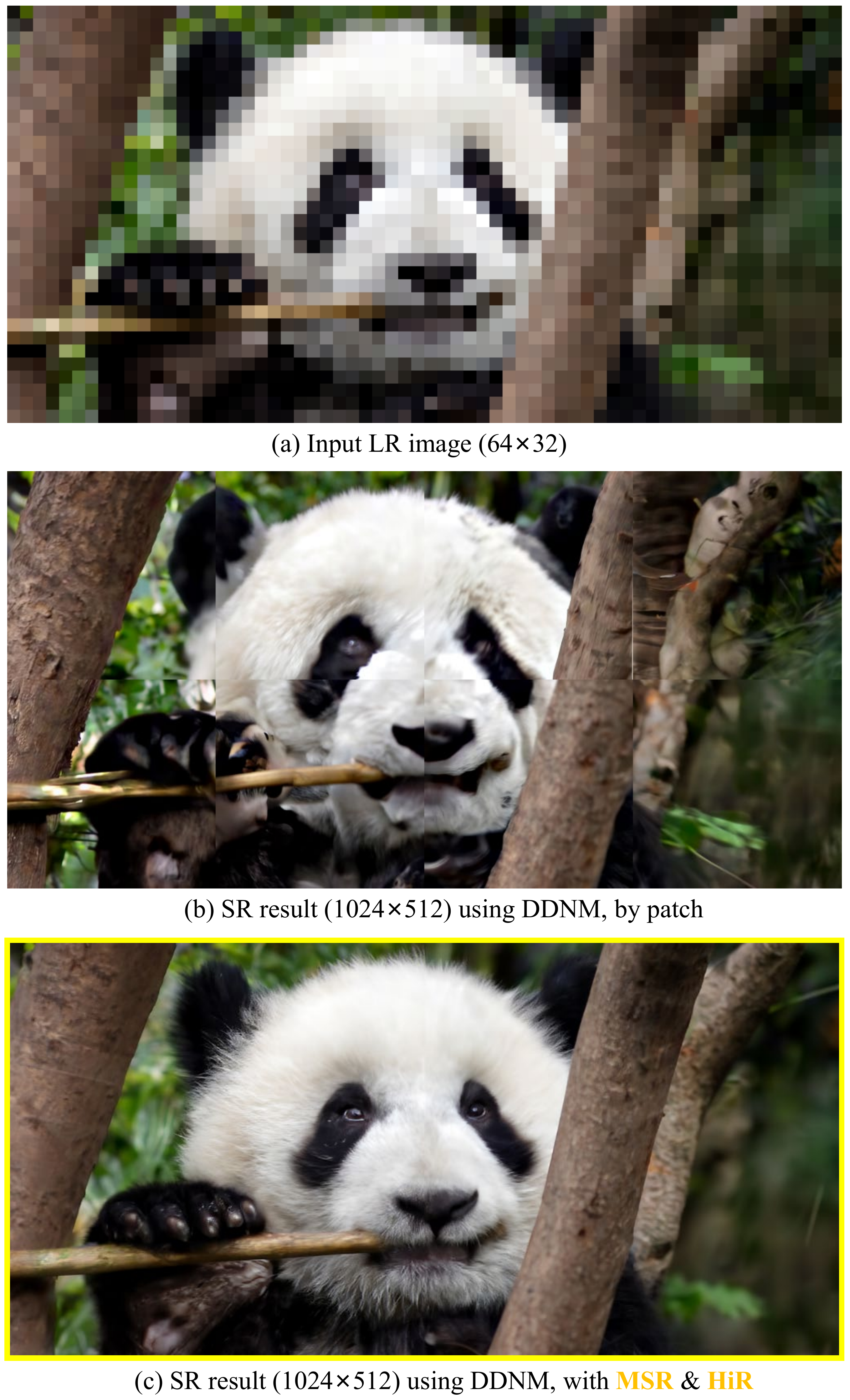}
  \caption{Example of 16$\times$ Super-Resolution (SR) that brings a 64$\times$32 Low-Resolution (LR) image into 1024$\times$512 SR results. (b) Simply dividing the result into eight 256$\times$256 patches and using DDNM \cite{wang2022zero} to solve them independently will get poor results, because it neither considers the global semantics nor the boundary information of adjacent patches. (c) We propose Mask-Shift Restoration (MSR) to solve the boundary artifacts and Hierarchical Restoration (HiR) to address the lack of global semantics.}
  \vspace{-0.5cm}
\label{fig:1} 
\end{figure}

\section{Introduction}
Recent progress in diffusion models \cite{song2019generative,song2020score,ho2020denoising,song2021denoising,dhariwal2021diffusion,bao2021analytic} has enlightened a lot works in solving Image Restoration (IR) tasks \cite{wang2022zero, choi2021ilvr, song2021solving, kawar2022jpeg, kawar2022denoising, lugmayr2022repaint, song2023pseudoinverseguided, Chung_2022_CVPR, chung2022improving, chung2022diffusion, sr3, saharia2022palette, whang2022deblurring}. These diffusion-based IR methods can be roughly divided into supervised \cite{sr3, saharia2022palette, whang2022deblurring, liu20232} and zero-shot \cite{wang2022zero, choi2021ilvr, song2021solving, kawar2022jpeg, kawar2022denoising, song2023pseudoinverseguided, lugmayr2022repaint, Chung_2022_CVPR, chung2022improving, chung2022diffusion}. Among them, zero-shot methods have developed a new hot paradigm since they only need to use the pre-trained off-the-shelf diffusion model, and can directly handle various IR tasks without any finetuning. Since zero-shot methods are usually independent of the choice of Diffusion Models, they can achieve better performance once a more powerful Diffusion Model is available. In this paper, we focus on zero-shot methods \cite{wang2022zero, choi2021ilvr, song2021solving, kawar2022jpeg, kawar2022denoising, song2023pseudoinverseguided, lugmayr2022repaint, Chung_2022_CVPR, chung2022improving, chung2022diffusion} which are concise, flexible, and in rapid progress. 

Existing diffusion-based IR methods mainly focus on IR problems with fixed output sizes. But in real-world applications, the desired output size may be arbitrary, depending on the user's demands. There are two main difficulties in applying these zero-shot IR methods to arbitrary output size: (1) The used diffusion models are usually pre-trained on fixed-size images, thus facing out-of-domain (OOD) issues when extending to arbitrary sizes; (2) The default network structure may not support arbitrary output size. The OOD issue can be solved by training the diffusion models with random cropped images. But the network structure constraint is hard to address. A common practice to bypass this constraint is to divide the input image into fixed-size patches and use the network to process each patch independently, then, concatenate the result patches as the final result, as shown in the middle of Fig.~\ref{fig:1}. However, this may lead to evident block artifacts and unreasonable restoration, because it neither considers the global semantics of all patches nor the local information of adjacent patches. 

We observe that the neighboring correlation is well considered in inpainting tasks in DDNM \cite{wang2022zero}, which inspired us to leave overlapped regions when dividing patches, then take the overlapped region as extra mask constraints when solving the following patches. We name this method Mask-Shift Restoration (MSR), which assures the coherence between patches and effectively eliminates boundary artifacts. 

To further alleviate the OOD problem, we propose to first restore the result at a small size, then use the small result as a global prior for the final result. We name this method Hierarchical Restoration (HiR). Note that both MSR and HiR perfectly fit the zero-shot properties, and can be flexibly combined. The bottom of Fig.~\ref{fig:1} shows the result using both MSR and HiR based on DDNM. From the perspective of Range-Null space Decomposition (RND), MSR and HiR are essentially adding extra linear constraints to the given inverse problem. This property makes it perfectly suitable for DDNM, which is exactly built on the principle of RND.

Our contribution includes:
\vspace{-0.2cm}
\begin{itemize}
    \item[1.] We propose Mask-Shift Restoration (MSR), a simple but effective method to eliminate boundary artifacts when processing a large image in patches.
    \vspace{-0.2cm}
    \item[1.] We propose Hierarchical Restoration (HiR) to alleviate the out-of-domain problem and the lack of global semantics when processing a large image in patches.
    \vspace{-0.2cm}
    \item[2.] We provide typical pipelines for using MSR and HiR for diverse applications, including but not limited to image generation, super-resolution, colorization, inpainting, and denoising. It is worth noting that our proposed methods are parameter-free and training-free, and can be applied to diverse diffusion models and zero-shot restoration methods. 
\end{itemize}

\section{Preliminaries}
\subsection{Diffusion Models}
Diffusion models have diverse interpretations \cite{song2020score, song2021denoising, bao2021analytic, lu2022dpm,liu2022flow}, but in this paper, we put aside the mathematical meaning and introduce the diffusion model in the most concise and general way.
Diffusion models \cite{song2019generative,song2020score,ho2020denoising,song2021denoising,dhariwal2021diffusion,bao2021analytic} define a $T$-step forward process and a $T$-step reverse process. The forward process adds random noise to data, while the reverse process constructs desired data samples from the noise. Specifically, the forward process yields a noisy image $\mathbf{x}_{t}$ from a clean image $\mathbf{x}_{0}$:
\begin{equation}
\mathbf{x}_{t}=a_{t}\mathbf{x}_{0}+\sigma_t\boldsymbol{\epsilon},\quad \boldsymbol{\epsilon}\sim \mathcal{N}(0,\mathbf{I})
    \label{eq:diffusion forward}
\end{equation}
where $t\sim\{0, ..., T\}$, $a_{t}$ and $\sigma_t$ are predefined scale factors, $\mathcal{N}$ represents the Gaussian distribution.

The core of the reverse process is estimating the clean image $\mathbf{x}_{0}$ from the noisy image $\mathbf{x}_{t}$:
\begin{equation}
\mathbf{x}_{0|t}=\frac{1}{a_{t}}(\mathbf{x}_{t}-\sigma_t\boldsymbol{\epsilon}_t)
    \label{eq:diffusion reverse}
\end{equation}
which is a reverse of Eq.~\ref{eq:diffusion forward}, with $\boldsymbol{\epsilon}_t$ denotes the estimation of noise $\boldsymbol{\epsilon}$ and $\mathbf{x}_{0|t}$ represents the estimation of $\mathbf{x}_{0}$ at time step $t$. Typically, a denoiser $\mathcal{Z}_{\boldsymbol{\theta}}$ is used to yield $\boldsymbol{\epsilon}_t$:
\begin{equation}
\boldsymbol{\epsilon}_t=\mathcal{Z}_{\boldsymbol{\theta}}(\mathbf{x}_{t}, t)
    \label{eq:predict noise}
\end{equation}
Then we can use Eq.~\ref{eq:diffusion forward} to generate the previous state $\mathbf{x}_{t-1}$, with $\mathbf{x}_{0|t}$ as the estimation of $\mathbf{x}_{0}$:
\begin{equation}
\mathbf{x}_{t-1}=a_{t-1}\mathbf{x}_{0|t}+\sigma_{t-1}\boldsymbol{\epsilon},\quad \boldsymbol{\epsilon}\sim \mathcal{N}(0,\mathbf{I})
    \label{eq:diffusion generation}
\end{equation}

With the above formulations, one can generate a clean image $\mathbf{x}_{0}$ from a random noise $\mathbf{x}_{T}$$\sim$$\mathcal{N}(\mathbf{0},\mathbf{I})$ by iterating Eq.~\ref{eq:diffusion reverse} and Eq.~\ref{eq:diffusion generation} while deceasing $t$ from $T$ to 0.

Such a reverse process is the simplest form. Further, for Eq.~\ref{eq:diffusion generation}, we can interpolate the newly added noise $\boldsymbol{\epsilon}$ with the estimated previous noise $\boldsymbol{\epsilon}_t$ under the premise of invariant total variance:
\begin{equation}
\mathbf{x}_{t-1}=a_{t-1}\mathbf{x}_{0|t}+\sigma_{t-1}(\eta_t\boldsymbol{\epsilon}+\sqrt{1-\eta_t^2}\boldsymbol{\epsilon}_t),\quad \boldsymbol{\epsilon}\sim \mathcal{N}(0,\mathbf{I})
    \label{eq:diffusion general generation}
\end{equation}
where $\eta_t$ is an interpolation factor that controls the ratio of the newly introduced noise $\boldsymbol{\epsilon}$. Note that Eq.~\ref{eq:diffusion general generation} describes a general form of reverse sampling methods. The critical difference between different sampling methods is the setting of $\eta_t$. For DDIM \cite{song2021denoising}, $\eta_t$ is a time-independent scalar; For DDPM \cite{ho2020denoising} and Analytic-DPM \cite{bao2021analytic}, $\eta_t$ is a time-dependent function. 

To train the denoiser $\mathcal{Z}_{\boldsymbol{\theta}}$, one can randomly pick a clean image $\mathbf{x}_{0}$ from the dataset and pick a random time-step $t$ to yield a noisy image $\mathbf{x}_{t}$ using Eq.~\ref{eq:diffusion forward}. Then, update the network parameters $\boldsymbol{\theta}$ with the following gradient descent step \cite{ho2020denoising}, and repeat the whole process until converged.
\begin{equation}
\nabla_{\boldsymbol{\theta}}||\boldsymbol{\epsilon}-\mathcal{Z}_{\boldsymbol{\theta}}(\mathbf{x}_{t},t)||^{2}_{2}.
    \label{eq:7}
\end{equation}

\subsection{Denoising Diffusion Null-space Model (DDNM)}
Recent progress shows that pre-trained diffusion models can be used to solve linear inverse problems in a zero-shot manner \cite{wang2022zero, lugmayr2022repaint, choi2021ilvr, song2021solving, kawar2022denoising}, without extra training or optimization. DDNM \cite{wang2022zero} explains the nature of such methods.

DDNM starts with noise-free linear image inverse problems. Given a degraded image $\mathbf{y}=\mathbf{A}\mathbf{x}$ where $\mathbf{A}$ is a linear operator and $\mathbf{x}$ is the original image, image restoration aims at yielding a result $\hat{\mathbf{x}}$ that satisfies two constraints:
\begin{equation}
    \textit{Consistency}: \quad \mathbf{A}\hat{\mathbf{x}} \equiv \mathbf{y},\quad\quad\textit{Realness}: \quad \hat{\mathbf{x}} \sim q(\mathbf{x}),
    \label{eq:consistency}
\end{equation}
where $q(\mathbf{x})$ denotes the distribution of the GT images.

Such a problem has a general solution that analytically satisfies the \textit{Consistency} constraint:
\begin{equation}
    \hat{\mathbf{x}}=\mathbf{A^{\dagger}}\mathbf{y} + (\mathbf{I} - \mathbf{A^{\dagger}}\mathbf{A})\mathbf{x}_r.
    \label{eq:rnd}
\end{equation}
where $\mathbf{A^{\dagger}}$ is the pseudo-inverse of $\mathbf{A}$ (satisfies $\mathbf{A}\mathbf{A^{\dagger}}\mathbf{A}\equiv\mathbf{A}$), and $\mathbf{x}_r$ is the unknown null-space variable to be solved. Note that Eq.~\ref{eq:rnd} originates from the Range-Null space Decomposition \cite{wang2022zero,wang2022gan,chen2020deep}. Another interpretation is that $\mathbf{A^{\dagger}}\mathbf{y}$ can be seen as a \textcolor{blue}{special solution} of $\mathbf{A}\mathbf{x}=\mathbf{y}$ since $\mathbf{A}\mathbf{A^{\dagger}}\mathbf{y}\equiv\mathbf{A}\mathbf{A^{\dagger}}\mathbf{A}\mathbf{x}\equiv\mathbf{A}\mathbf{x}\equiv\mathbf{y}$; and $(\mathbf{I} - \mathbf{A^{\dagger}}\mathbf{A})\mathbf{x}_r$ can be seen as a \textcolor{blue}{general solution} of $\mathbf{A}\mathbf{x}=\mathbf{0}$ since $\mathbf{A}(\mathbf{I} - \mathbf{A^{\dagger}}\mathbf{A})\mathbf{x}_r\equiv(\mathbf{A}-\mathbf{A})\mathbf{x}\equiv\mathbf{0}$ holds whatever $\mathbf{x}_r$ is.

To conclude, Eq.~\ref{eq:rnd} defined a solution that analytically satisfies the Consistency constraint but needs to find proper null-space variable $\mathbf{x}_r$ to meet the Realness constraint. As we will get to later, the methods proposed in this paper heavily rely on the use of Eq.~\ref{eq:rnd}.

\begin{figure}[t]
  \centering
  \includegraphics[width=1\linewidth]{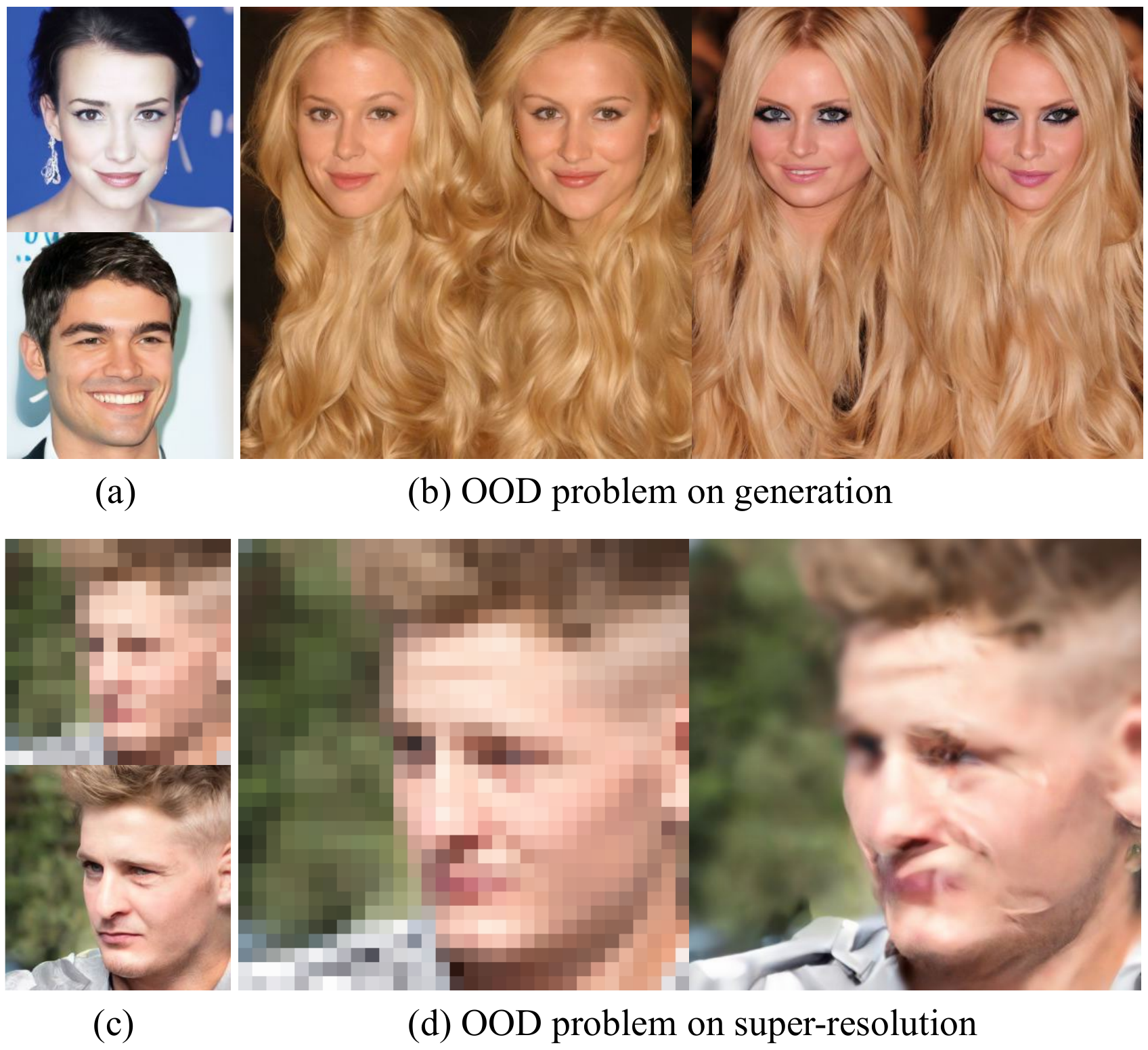}
  \caption{Out-Of-Domain (OOD) problem. (a) 256$\times$256 images generated by diffusion model trained on aligned 256$\times$256 CelebA dataset. (b) 512$\times$512 images generated by the same diffusion model. We can see that the model can not generate bigger faces even enforce to generate a 512$\times$512 image. (c) Applying the same diffusion model to DDNM \cite{wang2022zero} for 16$\times$ SR task yields good results of size 256$\times$256.  (d) Applying the same diffusion model to DDNM for 16$\times$ SR task yields terrible results of size 512$\times$512. This is caused by the OOD problem.}
\label{fig:2} 
\end{figure}

In DDNM \cite{wang2022zero}, the critical step using diffusion models for inverse problems is taking each estimation $\mathbf{x}_{0|t}$ as the null-space variable $\mathbf{x}_r$ in Eq.~\ref{eq:rnd}:
\begin{equation}
    \textcolor{blue}{\hat{\mathbf{x}}_{0|t}}=\mathbf{A^{\dagger}}\mathbf{y} + (\mathbf{I} - \mathbf{A^{\dagger}}\mathbf{A})\mathbf{x}_{0|t}.
    \label{eq:ddnm rnd}
\end{equation}
then use this consistent result $\textcolor{blue}{\hat{\mathbf{x}}_{0|t}}$ for subsequent sampling:
\begin{equation}
\mathbf{x}_{t-1}=a_{t-1}\textcolor{blue}{\hat{\mathbf{x}}_{0|t}}+\sigma_{t-1}(\eta_t\boldsymbol{\epsilon}+\sqrt{1-\eta_t^2}\boldsymbol{\epsilon}_t),\quad \boldsymbol{\epsilon}\sim \mathcal{N}(0,\mathbf{I})
    \label{eq:ddnm reverse}
\end{equation}
Algo.~\ref{alg:ddnm} shows the whole process of DDNM. See Appendix \ref{app: ddnm noisy} for DDNM with noisy situations. 
\begin{algorithm}[H]
\footnotesize
\caption{Sampling process of DDNM}
\label{alg:ddnm}
\begin{algorithmic}[1]
    \State $\mathbf{x}_{T}\sim\mathcal{N}(\mathbf{0},\mathbf{I})$
    \For{$t = T, ..., 1$}
         \State $\boldsymbol{\epsilon}_t=\mathcal{Z}_{\boldsymbol{\theta}}(\mathbf{x}_{t}, t)$
        \State $\mathbf{x}_{0|t}=\frac{1}{a_{t}}(\mathbf{x}_{t}-\sigma_t\boldsymbol{\epsilon}_t)$
        \State $\textcolor{blue}{\hat{\mathbf{x}}_{0|t}} = \mathbf{A}^{\dagger}\mathbf{y} + (\mathbf{I} - \mathbf{A}^{\dagger}\mathbf{A})\mathbf{x}_{0|t}$
        \State $\mathbf{x}_{t-1}=a_{t-1}\textcolor{blue}{\hat{\mathbf{x}}_{0|t}}+\sigma_{t-1}(\eta_t\boldsymbol{\epsilon}+\sqrt{1-\eta_t^2}\boldsymbol{\epsilon}_t),\quad \boldsymbol{\epsilon}\sim \mathcal{N}(0,\mathbf{I})$
    \EndFor
    \State \textbf{return} $\mathbf{x}_{0}$
\end{algorithmic}
\end{algorithm}
\begin{algorithm}[H]
\footnotesize
\caption{Mask-Shift Restoration, based on DDNM}
\label{alg:msr}
\textbf{Additional Requirement}: The already restored region $\dot{\mathbf{x}}_{0}$ and the corresponding mask $\mathbf{A_{m}}$.
\begin{algorithmic}[1]

    \State $\mathbf{x}_{T}\sim\mathcal{N}(\mathbf{0},\mathbf{I})$
    \For{$t = T, ..., 1$}
        \State $\boldsymbol{\epsilon}_t=\mathcal{Z}_{\boldsymbol{\theta}}(\mathbf{x}_{t}, t)$
        \State $\mathbf{x}_{0|t}=\frac{1}{a_{t}}(\mathbf{x}_{t}-\sigma_t\boldsymbol{\epsilon}_t)$
        \State $\hat{\mathbf{x}}_{0|t}=\mathbf{A^{\dagger}}\mathbf{y} + (\mathbf{I} - \mathbf{A^{\dagger}}\mathbf{A})\mathbf{x}_{0|t}$
        \State $\textcolor{blue}{\bar{\mathbf{x}}_{0|t}=\mathbf{A}_m\dot{\mathbf{x}}_{0} + (\mathbf{I} - \mathbf{A}_m)\hat{\mathbf{x}}_{0|t}}$
        \State $\mathbf{x}_{t-1}=a_{t-1}\textcolor{blue}{\Bar{\mathbf{x}}_{0|t}}+\sigma_{t-1}(\eta_t\boldsymbol{\epsilon}+\sqrt{1-\eta_t^2}\boldsymbol{\epsilon}_t),\quad \boldsymbol{\epsilon}\sim \mathcal{N}(0,\mathbf{I})$
    \EndFor
    \State \textbf{return} $\mathbf{x}_{0}$
\end{algorithmic}
\end{algorithm}
\vspace{-0.2cm}
\begin{algorithm}[H]
\footnotesize
\caption{Hierarchical Restoration, based on DDNM}
\label{alg:hir}
\textbf{Additional Requirement}: The low-resolution result $\ddot{\mathbf{x}}_{0}$ and the corresponding downsampler $\mathbf{A_{sr}}$ and its pseudo-inverse $\mathbf{A^{\dagger}_{sr}}$. The already restored region $\dot{\mathbf{x}}_{0}$ and the corresponding mask $\mathbf{A_{m}}$. 
\begin{algorithmic}[1]
    \State $\mathbf{x}_{T}\sim\mathcal{N}(\mathbf{0},\mathbf{I})$
    \For{$t = T, ..., 1$}
        \State $\boldsymbol{\epsilon}_t=\mathcal{Z}_{\boldsymbol{\theta}}(\mathbf{x}_{t}, t)$
        \State $\mathbf{x}_{0|t}=\frac{1}{a_{t}}(\mathbf{x}_{t}-\sigma_t\boldsymbol{\epsilon}_t)$
        \State$\textcolor{blue}{\tilde{\mathbf{x}}_{0|t}=\mathbf{A_{sr}^{\dagger}}\ddot{\mathbf{x}}_{0} + (\mathbf{I} - \mathbf{A_{sr}^{\dagger}}\mathbf{A_{sr}})\mathbf{x}_{0|t}}$
        \State $\hat{\mathbf{x}}_{0|t}=\mathbf{A^{\dagger}}\mathbf{y} + (\mathbf{I} - \mathbf{A^{\dagger}}\mathbf{A})\textcolor{blue}{\tilde{\mathbf{x}}_{0|t}}$
        \State $\textcolor{blue}{\bar{\mathbf{x}}_{0|t}=\mathbf{A}_m\dot{\mathbf{x}}_{0} + (\mathbf{I} - \mathbf{A}_m)\hat{\mathbf{x}}_{0|t}}$
        \State $\mathbf{x}_{t-1}=a_{t-1}\textcolor{blue}{\Bar{\mathbf{x}}_{0|t}}+\sigma_{t-1}(\eta_t\boldsymbol{\epsilon}+\sqrt{1-\eta_t^2}\boldsymbol{\epsilon}_t),\quad \boldsymbol{\epsilon}\sim \mathcal{N}(0,\mathbf{I})$
    \EndFor
    \State \textbf{return} $\mathbf{x}_{0}$
\end{algorithmic}
\end{algorithm}

\vspace{-1cm}

\begin{figure*}[t]
  \centering
  \includegraphics[width=1\linewidth]{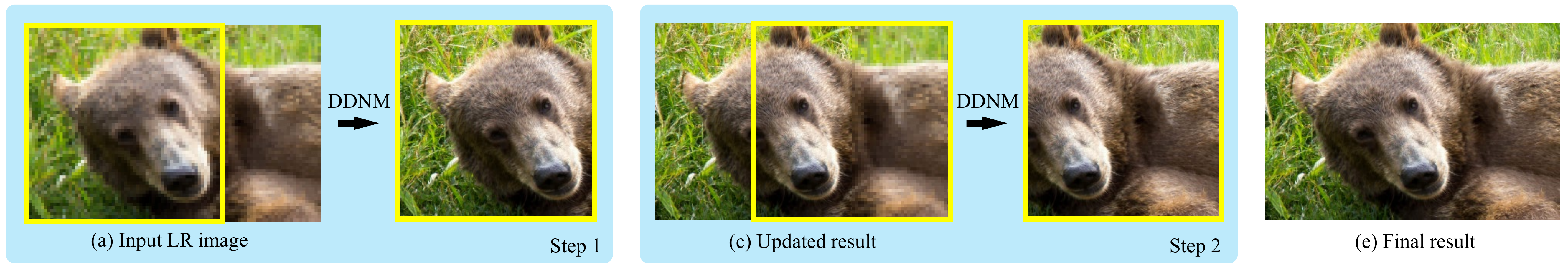}
  \caption{Example of Mask-Shift Restoration for 4$\times$ SR. Given an input LR image (a) with a non-square size, we first use DDNM to SR the first square patch and update the result (Step 1). Then, we shift the patch right, leaving some overlapped regions with the previous patch. Since the overlapped region is already restored, we set them as fixed and only solve the rest region (Step 2). To this end, we need an extra inpainting (mask) constraint, which is perfectly suitable for zero-shot methods like DDNM to handle. \textbf{Zoom in for the best view}}
\label{fig:3} 
\end{figure*}

\section{Method}
We have introduced the basic principles of the diffusion model and DDNM. We can see that the limitation of the image processing size lies in the denoiser. Usually, the denoiser is pre-trained on fixed-size images. How do we use such pre-trained denoisers for unlimited-size image restoration? In the following part, we propose two methods to achieve this goal, both inherit the zero-shot property.

\subsection{Process as a Whole Image}
Typical diffusion models \cite{song2019generative,song2020score,ho2020denoising,song2021denoising,dhariwal2021diffusion,bao2021analytic} use U-Net structures \cite{ronneberger2015u} as the denoiser backbone. Theoretically, U-Net is a convolutional network and thus supports scalable input size.

Hence a simple solution is to directly change the model processing size. A similar approach has been widely adopted by Stable Diffusion \cite{rombach2022high} for flexible generated size. Despite supporting flexible input size, the denoiser trained on fixed image size may face Out-Of-Domain (OOD) problem when applied to other image sizes. As shown in Fig.~\ref{fig:2}, a diffusion model trained on CelebA 256$\times$256 fails to generate desired 512$\times$512 face images. One way to solve the OOD issue is to train the 256$\times$256 denoiser with a random cropped dataset, rather than an aligned one. Interestingly, ImageNet and LAION-5B happen to be non-aligned datasets, and hence suffer relatively minor OOD issues.

\subsection{Process as Patches}
Directly changing the model
processing size may work, but it still has the following limitations: (1) It may yield bad results when facing OOD problems, as shown in Fig.~\ref{fig:2}(b). (2) It still has limitations on image size, e.g., divisible by 32; (3) Large sizes, e.g., 1024$\times$1024, may cause unaffordable memory consumption; (4) The classifier guidance \cite{dhariwal2021diffusion} can not be applied since it is usually designed for fixed input sizes; (5) Other potential network backbones \cite{peebles2022scalable} may not support flexible processing size.  

How to use diffusion models with fixed processing sizes to solve arbitrary image sizes? A simple solution is dividing the input image $\mathbf{y}$ into patches, solving each patch independently, then concatenating the results. But this may cause evident boundary artifacts, as shown in the middle of Fig.~\ref{fig:1}. This is because each patch is solved independently and their connection is not considered.

\begin{figure*}[t]
  \centering
  \includegraphics[width=1\linewidth]{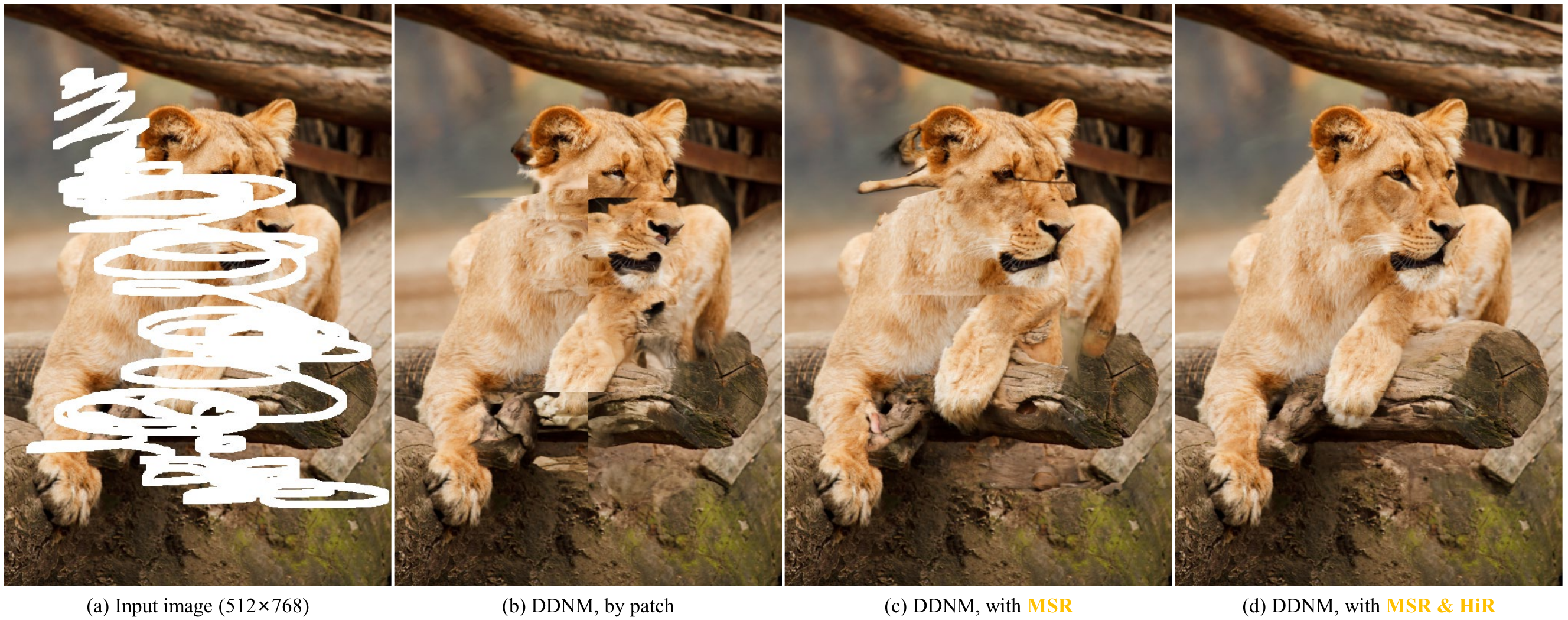}
  \caption{Comparison on large scale inpainting. (b) and (c) yields unreasonable results since the patch is too small to capture global semantic information. In contrast, (c) yields a decent result due to the use of Hierarchical Restoration (HiR). \textbf{Zoom in for the best view}.}
\label{fig:4} 
\end{figure*}

\begin{figure*}[t]
  \centering
  \includegraphics[width=1\linewidth]{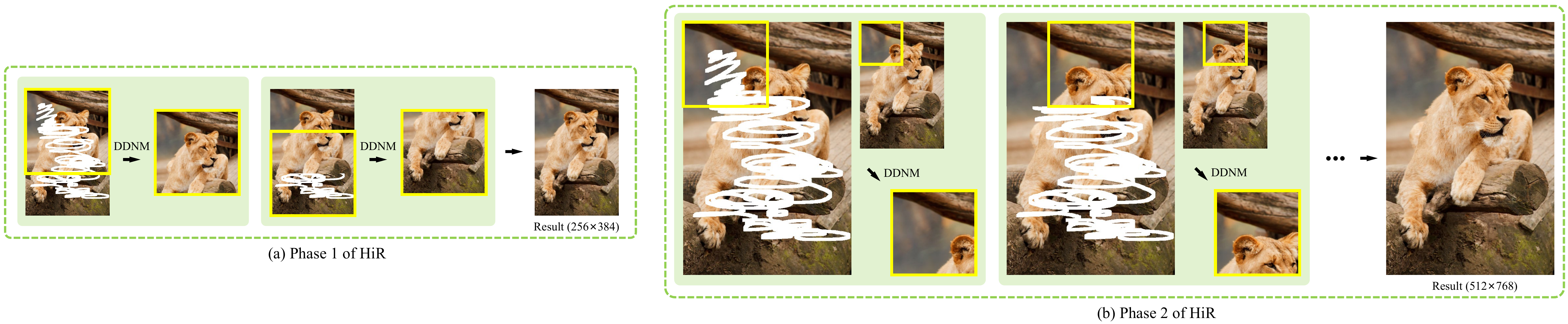}
  \caption{Example of Hierarchical Restoration for Inpainting. (a) We first do a 2$\times$ downsampling for Fig.~\ref{fig:4}(a) and use MSR to restore a small result. (b) Then we use this small result as extra low-frequency guidance, and use MSR at the original size to yield the final result.}
\label{fig:5} 
\end{figure*}

\subsection{Mask-Shift Restoration}
Among the many image restoration tasks, inpainting is the typical one that considers the connection between the masked and unmasked region. Zero-shot methods like DDNM \cite{wang2022zero} and RePaint \cite{lugmayr2022repaint} show good performance in solving inpainting.

Our insight is that we can leave overlapped regions when dividing patches, then take these overlapped regions as an extra constraint when solving the following patches. The neat thing is that this constraint can be integrated into existing zero-shot methods \cite{wang2022zero, choi2021ilvr, song2021solving, kawar2022jpeg, kawar2022denoising, song2023pseudoinverseguided, lugmayr2022repaint, Chung_2022_CVPR, chung2022improving, chung2022diffusion}, with just one extra line of code!

Let's take a 4$\times$SR task for example, as shown in Fig.~\ref{fig:3}. Given an input image $\mathbf{y}^{full}$ with size 64$\times$96, our aim is to get an SR result with size 256$\times$384. Here we set the degradation operator $\mathbf{A}$ as the average-pooling downsampler, and its pseudo-inverse $\mathbf{A}^{\dagger}$ as the replication upsampler  \cite{wang2022gan}. Fig.~\ref{fig:3}(a) shows the result of $\mathbf{A}^{\dagger}\mathbf{y}^{full}$. We first divide $\mathbf{A}^{\dagger}\mathbf{y}^{full}$ into two square patches $\mathbf{A}^{\dagger}\dot{\mathbf{y}}$ and $\mathbf{A}^{\dagger}\mathbf{y}$ of size 256$\times$256. Note that $\mathbf{A}^{\dagger}\dot{\mathbf{y}}$ and $\mathbf{A}^{\dagger}\mathbf{y}$ has an overlap of size 256$\times$128.

We first use default DDNM to process $\mathbf{A}^{\dagger}\dot{\mathbf{y}}$ and get the SR result $\dot{\mathbf{x}}_{0}$ (Step 1 in Fig.~\ref{fig:3}). Note that $\mathbf{A}^{\dagger}\dot{\mathbf{y}}$ and $\mathbf{A}^{\dagger}\mathbf{y}$ has an overlap of size 256$\times$128, and this overlapped region is already restored in $\dot{\mathbf{x}}_{0}$. So when we use DDNM to solve $\mathbf{A}^{\dagger}\mathbf{y}$, we can take the restored overlapped region as a known part in an inpainting setting (Step 2 in Fig.~\ref{fig:3}). Specifically, we insert an extra inpainting constraint behind Eq.~\ref{eq:ddnm rnd} in DDNM: 
\begin{equation}
\bar{\mathbf{x}}_{0|t}=\mathbf{A}_{m}\dot{\mathbf{x}}_{0} + (\mathbf{I} - \mathbf{A}_{m})\hat{\mathbf{x}}_{0|t}.
    \label{eq: maskshift}
\end{equation}
where $\mathbf{A}_{m}$ denotes the mask operator for overlapped region between $\mathbf{A}^{\dagger}\dot{\mathbf{y}}$ and $\mathbf{A}^{\dagger}\mathbf{y}$. The whole algorithm is summarized in Algo.~\ref{alg:msr}, named as Mask-Shift Restoration (MSR).
 
As we can see from Fig.~\ref{fig:3}(c), the final result concatenated by the results of Step 1 and Step 2 does not show boundary artifacts. Similarly, we can iteratively use MSR to generate an unlimited-size image without boundary artifacts. Note that the overlapped region and the shifted direction can be arbitrary, and the supported task is also not limited to SR, but to all linear inverse problems.

\subsection{Hierarchical Restoration}
Though MSR assures local coherence, it owns a small receptive field when dealing with a large image. This may lead to a lack of grasp of global information, resulting in poor semantic information recovery. In Fig.~\ref{fig:4}(a) we show a masked image of size 512$\times$768, where any 256$\times$256 patch can not cover the whole semantic subject. Fig.~\ref{fig:4}(b) shows the result using MSR based on DDNM. Though with good local coherence, it yields unreasonable semantic structures.

To extend the receptive field for better semantic restoration, we propose Hierarchical Restoration (HiR). HiR consists of two phases: a semantic restoration phase and a texture restoration phase. 

Take Fig.~\ref{fig:4}(a) for example. For the semantic restoration phase, we first undergo a 2$\times$ downsample to convert the 512$\times$768 input into a 256$\times$384 one, where a 256$\times$256 patch can cover the whole semantic subject. Then we use MSR based on DDNM to get a 256$\times$384 inpainting result $\ddot{\mathbf{x}}_{0}$, as shown in Fig.~\ref{fig:5}(a). This result is semantically reasonable and can be used as a low-frequency reference. For the texture restoration phase (Fig.~\ref{fig:5}(b)), we add an extra low-frequency constraint before Eq.~\ref{eq:ddnm rnd}:  
\begin{equation}
    \textcolor{blue}{\tilde{\mathbf{x}}_{0|t}}=\mathbf{A_{sr}^{\dagger}}\ddot{\mathbf{x}}_{0} + (\mathbf{I} - \mathbf{A_{sr}^{\dagger}}\mathbf{A_{sr}})\mathbf{x}_{0|t}.
    \label{eq:HiR}
\end{equation}
where $\mathbf{A_{sr}}$ and $\mathbf{A_{sr}^{\dagger}}$ represent the average-pooling downsampler and its pseudo-inverse upsampler \cite{wang2022gan}, respectively. Algo.~\ref{alg:hir} shows the whole algorithm of the second phase of HiR.

As we can see from Fig.~\ref{fig:4}(d), the use of HiR significantly improves semantic correctness. Note that the HiR is not limited to inpainting tasks, but is also useful for large-scale SR (Fig.~\ref{fig:1}(c)) and colorization (Fig.~\ref{fig:7}), etc.

\subsection{Flexible Pipeline for Applications}
Mask-Shift Restoration (MSR) can be seen as a general patch connection technology, and Hierarchical Restoration (HiR) can be seen as a general method to improve restoration quality. The essence of both MSR and HiR is to determine part of the information via prior knowledge to narrow the solution space. In this paper, we implement MSR and HiR via the Range-Null space Decomposition, which is concise, effective, and mathematically elegant. Besides, there remain other possible ways to implement MSR and HiR, e.g., adding extra loss into optimization-based methods such as DPS. Hence the proposed MSR and HiR can be also used for other diffusion-based zero-shot IR methods, e.g., ILVR\cite{choi2021ilvr}, RePaint\cite{lugmayr2022repaint}, and DPS\cite{chung2022diffusion}.

\section{Experiment}
In this section, we describe the configuration of the experiment in detail. All experiments use the denoiser pre-trained on ImageNet 256$\times$256, provided by guided-diffusion \cite{dhariwal2021diffusion}. We use the classifier guidance \cite{dhariwal2021diffusion} for sampling. Besides, the time-travel sampling \cite{wang2022zero, lugmayr2022repaint} is also used to improve the generative quality.

\begin{figure}[t]
  \centering
  \includegraphics[width=1\linewidth]{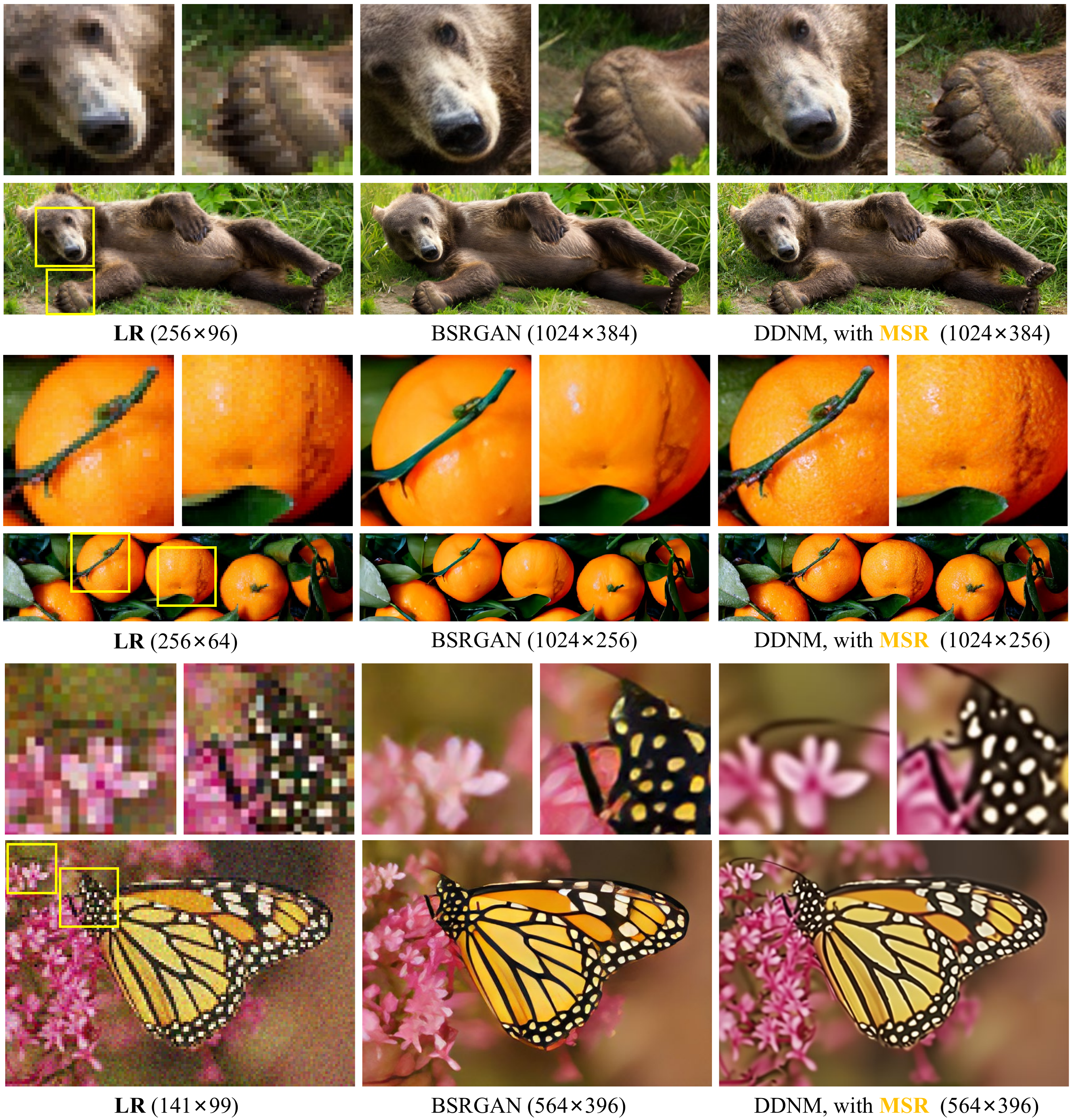}
  \caption{Experiment on noisy 4$\times$ SR. Compared with BSRGAN \cite{zhang2021designing}, a supervised IR method, we can see that our method performs better in both realness and consistency. Due to the use of RND \cite{wang2022zero,wang2022gan,chen2020deep}, our method can faithfully inherit the correct color and structure information in LR, while BSRGAN \cite{zhang2021designing} fails (see the results of butterfly). \textbf{Zoom in for the best view} }
\label{fig:8} 
\end{figure}

\begin{figure}[t]
  \centering
  \includegraphics[width=1\linewidth]{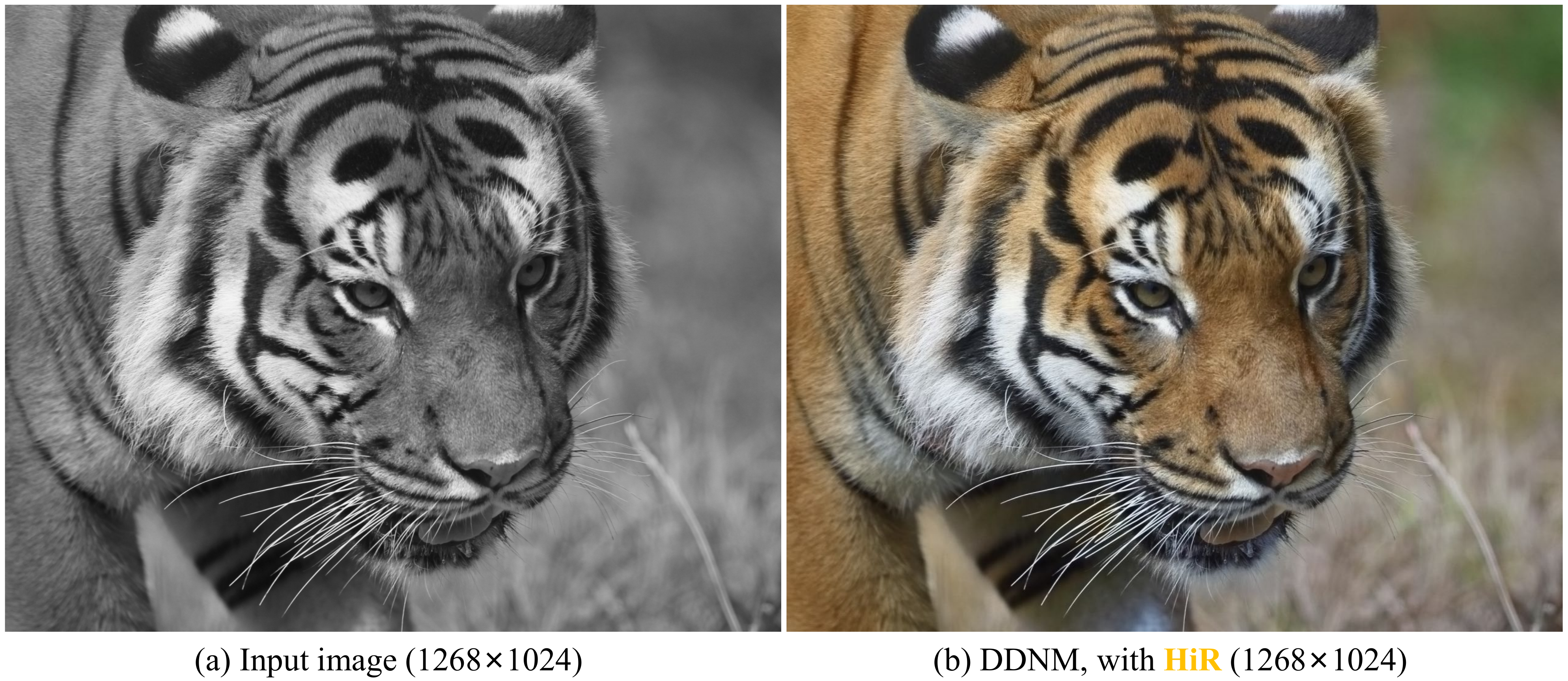}
  \caption{Colorization using HiR. \textbf{Zoom in for the best view}}
\label{fig:7} 
\end{figure}

Given a desired result size, we divide it into patches from left to right, top to bottom. Each patch has a size 256$\times$256 and has overlaps of 128 pixels with its neighbor patch, except for the boundary case. We solve the first patch using the original DDNM and solve the following patches in sequence (left to right, top to bottom) using MSR based on DDNM. Fig.~\ref{fig:3} shows the results on 4$\times$ SR, with $T=100$, time-travel length \cite{wang2022zero} $l=10$, repeat times $r=3$. In Fig.~\ref{fig:8}, we present qualitative comparisons between BSRGAN \cite{zhang2021designing} and MSR-based DDNM. We experiment on 4$\times$ SR and noisy 4$\times$ SR of different sizes, where MSR-based DDNM uses $T=250$, $l=10$, and $r=3$.  For Fig.~\ref{fig:1}(c), Fig.~\ref{fig:4}(d), and Fig.~\ref{fig:7} we use HiR based on DDNM.

\section{Related Work}
\textbf{Range-Null space Decomposition (RND)} \cite{rnd} is a concept in linear algebra. When applied to linear inverse problems, RND explicitly defines the upper limit of recoverable information. Chen et al. \cite{chen2020deep} introduce RND into image inverse problems, and propose learning the range and null space respectively. Wang et al. \cite{wang2022gan} propose using GAN Prior to learn the Null-space and propose using average-pooling and its pseudo-inverse as a general tool for SR tasks. In DDNM \cite{wang2022zero}, the authors propose using diffusion sampling to learn the Null-space and propose several practical operators for diverse applications.

\textbf{Diffusion-based Zero-Shot Image Restoration Methods} can be roughly divided into RND-based \cite{wang2022zero, choi2021ilvr, song2021solving, kawar2022jpeg, kawar2022denoising, song2023pseudoinverseguided, lugmayr2022repaint} and optimization-based \cite{Chung_2022_CVPR, chung2022improving, chung2022diffusion}. The essence of these two branches lies in modifying only the sampling process while keeping the network unchanged. Specifically, they modify the intermediate image $\mathbf{x}_{0|t}$ or its noisy version $\mathbf{x}_{t}$. For a given input and a certain degradation operator, RND-based methods use RND to explicitly assure the data consistency of $\mathbf{x}_{0|t}$ or $\mathbf{x}_{t}$, while optimization-based methods optimize $\mathbf{x}_{0|t}$ or $\mathbf{x}_{t}$ toward the data consistency. Generally speaking, the RND-based methods perform better in linear inverse problems but can not solve non-linear problems. The optimization-based methods cost more on memory and inference time but can support any differentiable operator, even as a complex network \cite{bansal2023universal}.

\section{Limitations \& Discussions}
Zero-shot IR methods \cite{wang2022zero, choi2021ilvr, song2021solving, kawar2022jpeg, kawar2022denoising, song2023pseudoinverseguided, lugmayr2022repaint, Chung_2022_CVPR, chung2022improving, chung2022diffusion} using diffusion models certainly open up a promising new direction for IR problems. The method proposed in this paper further enables those methods to support unlimited image size. However, there remain some limitations to be solved. Firstly, the calculation and time consumption are significantly more than those prevailing supervised methods. Secondly, the ceiling of performance depends on the pre-trained diffusion models. It may yield more interesting applications if applying our method to models like Imagen \cite{saharia2022photorealistic}, but they are not open-sourced yet. On the other hand, wildly used models like Stable Diffusion \cite{rombach2022high} are based on latent space, which makes it difficult to apply zero-shot methods. Thirdly, the degradation operator is explicitly needed, which makes it difficult for tasks like rain and haze removal. 

Another interesting observation is that MSR can be seen as a general image connection method, where we can use different models to restore special crops, e.g., use face restoration models \cite{gu2022vqfr, gfpgan, wang2022gan, panini, hu2023dear} for face crops, then fuse them with the background using MSR to avoid boundary artifacts.

{\small
\bibliographystyle{ieee_fullname}
\bibliography{egbib}

\begin{thebibliography}{10}\itemsep=-1pt

\bibitem{bansal2023universal}
Arpit Bansal, Hong-Min Chu, Avi Schwarzschild, Soumyadip Sengupta, Micah
  Goldblum, Jonas Geiping, and Tom Goldstein.
\newblock Universal guidance for diffusion models.
\newblock {\em arXiv preprint arXiv:2302.07121}, 2023.

\bibitem{bao2021analytic}
Fan Bao, Chongxuan Li, Jun Zhu, and Bo Zhang.
\newblock Analytic-dpm: an analytic estimate of the optimal reverse variance in
  diffusion probabilistic models.
\newblock In {\em International Conference on Learning Representations (ICLR)},
  2022.

\bibitem{chen2020deep}
Dongdong Chen and Mike~E Davies.
\newblock Deep decomposition learning for inverse imaging problems.
\newblock In {\em European Conference on Computer Vision (ECCV)}. Springer,
  2020.

\bibitem{choi2021ilvr}
Jooyoung Choi, Sungwon Kim, Yonghyun Jeong, Youngjune Gwon, and Sungroh Yoon.
\newblock Ilvr: Conditioning method for denoising diffusion probabilistic
  models.
\newblock In {\em Proceedings of the IEEE/CVF International Conference on
  Computer Vision (ICCV)}, 2021.

\bibitem{chung2022diffusion}
Hyungjin Chung, Jeongsol Kim, Michael~T Mccann, Marc~L Klasky, and Jong~Chul
  Ye.
\newblock Diffusion posterior sampling for general noisy inverse problems.
\newblock {\em International Conference on Learning Representations (ICLR)},
  2023.

\bibitem{Chung_2022_CVPR}
Hyungjin Chung, Byeongsu Sim, and Jong~Chul Ye.
\newblock Come-closer-diffuse-faster: Accelerating conditional diffusion models
  for inverse problems through stochastic contraction.
\newblock In {\em Proceedings of the IEEE/CVF Conference on Computer Vision and
  Pattern Recognition (CVPR)}, 2022.

\bibitem{chung2022improving}
Hyungjin Chung, Byeongsu Sim, and Jong~Chul Ye.
\newblock Improving diffusion models for inverse problems using manifold
  constraints.
\newblock In {\em Advances in Neural Information Processing Systems (NeurIPS)},
  2022.

\bibitem{dhariwal2021diffusion}
Prafulla Dhariwal and Alexander Nichol.
\newblock Diffusion models beat gans on image synthesis.
\newblock {\em Advances in Neural Information Processing Systems (NeurIPS)},
  34, 2021.

\bibitem{gu2022vqfr}
Yuchao Gu, Xintao Wang, Liangbin Xie, Chao Dong, Gen Li, Ying Shan, and
  Ming-Ming Cheng.
\newblock Vqfr: Blind face restoration with vector-quantized dictionary and
  parallel decoder.
\newblock In {\em Computer Vision--ECCV 2022: 17th European Conference, Tel
  Aviv, Israel, October 23--27, 2022, Proceedings, Part XVIII}, pages 126--143.
  Springer, 2022.

\bibitem{ho2020denoising}
Jonathan Ho, Ajay Jain, and Pieter Abbeel.
\newblock Denoising diffusion probabilistic models.
\newblock {\em Advances in Neural Information Processing Systems (NeurIPS)},
  33, 2020.

\bibitem{hu2023dear}
Yujie Hu, Yinhuai Wang, and Jian Zhang.
\newblock Dear-gan: Degradation-aware face restoration with gan prior.
\newblock {\em IEEE Transactions on Circuits and Systems for Video Technology},
  2023.

\bibitem{kawar2022denoising}
Bahjat Kawar, Michael Elad, Stefano Ermon, and Jiaming Song.
\newblock Denoising diffusion restoration models.
\newblock In {\em Advances in Neural Information Processing Systems (NeurIPS)},
  2022.

\bibitem{kawar2022jpeg}
Bahjat Kawar, Jiaming Song, Stefano Ermon, and Michael Elad.
\newblock Jpeg artifact correction using denoising diffusion restoration
  models.
\newblock In {\em Neural Information Processing Systems (NeurIPS) Workshop on
  Score-Based Methods}, 2022.

\bibitem{liu20232}
Guan-Horng Liu, Arash Vahdat, De-An Huang, Evangelos~A Theodorou, Weili Nie,
  and Anima Anandkumar.
\newblock I$^{2}$sb: Image-to-image schrodinger bridge.
\newblock {\em arXiv preprint arXiv:2302.05872}, 2023.

\bibitem{liu2022flow}
Xingchao Liu, Chengyue Gong, and Qiang Liu.
\newblock Flow straight and fast: Learning to generate and transfer data with
  rectified flow.
\newblock {\em International Conference on Learning Representations (ICLR)},
  2023.

\bibitem{lu2022dpm}
Cheng Lu, Yuhao Zhou, Fan Bao, Jianfei Chen, Chongxuan Li, and Jun Zhu.
\newblock Dpm-solver: A fast ode solver for diffusion probabilistic model
  sampling in around 10 steps.
\newblock {\em Advances in Neural Information Processing Systems (NeurIPS)},
  2022.

\bibitem{lugmayr2022repaint}
Andreas Lugmayr, Martin Danelljan, Andres Romero, Fisher Yu, Radu Timofte, and
  Luc Van~Gool.
\newblock Repaint: Inpainting using denoising diffusion probabilistic models.
\newblock In {\em Proceedings of the IEEE/CVF Conference on Computer Vision and
  Pattern Recognition (CVPR)}, 2022.

\bibitem{peebles2022scalable}
William Peebles and Saining Xie.
\newblock Scalable diffusion models with transformers.
\newblock {\em arXiv e-prints}, pages arXiv--2212, 2022.

\bibitem{rombach2022high}
Robin Rombach, Andreas Blattmann, Dominik Lorenz, Patrick Esser, and Bj{\"o}rn
  Ommer.
\newblock High-resolution image synthesis with latent diffusion models.
\newblock In {\em Proceedings of the IEEE/CVF Conference on Computer Vision and
  Pattern Recognition (CVPR)}, pages 10684--10695, 2022.

\bibitem{ronneberger2015u}
Olaf Ronneberger, Philipp Fischer, and Thomas Brox.
\newblock U-net: Convolutional networks for biomedical image segmentation.
\newblock In {\em International Conference on Medical Image Computing and
  Computer-Assisted Intervention (MICCAI)}, pages 234--241. Springer, 2015.

\bibitem{saharia2022palette}
Chitwan Saharia, William Chan, Huiwen Chang, Chris Lee, Jonathan Ho, Tim
  Salimans, David Fleet, and Mohammad Norouzi.
\newblock Palette: Image-to-image diffusion models.
\newblock In {\em ACM SIGGRAPH 2022 Conference Proceedings}, 2022.

\bibitem{saharia2022photorealistic}
Chitwan Saharia, William Chan, Saurabh Saxena, Lala Li, Jay Whang, Emily
  Denton, Seyed Kamyar~Seyed Ghasemipour, Burcu~Karagol Ayan, S~Sara Mahdavi,
  Rapha~Gontijo Lopes, et~al.
\newblock Photorealistic text-to-image diffusion models with deep language
  understanding.
\newblock {\em arXiv preprint arXiv:2205.11487}, 2022.

\bibitem{sr3}
Chitwan Saharia, Jonathan Ho, William Chan, Tim Salimans, David~J Fleet, and
  Mohammad Norouzi.
\newblock Image super-resolution via iterative refinement.
\newblock {\em IEEE Transactions on Pattern Analysis and Machine Intelligence},
  2022.

\bibitem{song2021denoising}
Jiaming Song, Chenlin Meng, and Stefano Ermon.
\newblock Denoising diffusion implicit models.
\newblock In {\em International Conference on Learning Representations (ICLR)},
  2021.

\bibitem{song2023pseudoinverseguided}
Jiaming Song, Arash Vahdat, Morteza Mardani, and Jan Kautz.
\newblock Pseudoinverse-guided diffusion models for inverse problems.
\newblock In {\em International Conference on Learning Representations (ICLR)},
  2023.

\bibitem{song2019generative}
Yang Song and Stefano Ermon.
\newblock Generative modeling by estimating gradients of the data distribution.
\newblock {\em Advances in Neural Information Processing Systems (NeurIPS)},
  32, 2019.

\bibitem{song2021solving}
Yang Song, Liyue Shen, Lei Xing, and Stefano Ermon.
\newblock Solving inverse problems in medical imaging with score-based
  generative models.
\newblock In {\em International Conference on Learning Representations (ICLR)},
  2021.

\bibitem{song2020score}
Yang Song, Jascha Sohl-Dickstein, Diederik~P Kingma, Abhishek Kumar, Stefano
  Ermon, and Ben Poole.
\newblock Score-based generative modeling through stochastic differential
  equations.
\newblock In {\em International Conference on Learning Representations (ICLR)},
  2020.

\bibitem{rnd}
Marco Taboga.
\newblock Range null-space decomposition.
\newblock In {\em Lectures on matrix algebra.
  https://www.statlect.com/matrix-algebra/range-null-space-decomposition.},
  2021.

\bibitem{gfpgan}
Xintao Wang, Yu Li, Honglun Zhang, and Ying Shan.
\newblock Towards real-world blind face restoration with generative facial
  prior.
\newblock In {\em Proceedings of the IEEE/CVF Conference on Computer Vision and
  Pattern Recognition (CVPR)}, 2021.

\bibitem{wang2022gan}
Yinhuai Wang, Yujie Hu, Jiwen Yu, and Jian Zhang.
\newblock Gan prior based null-space learning for consistent super-resolution.
\newblock {\em Proceedings of the AAAI Conference on Artificial Intelligence
  (AAAI)}, 2023.

\bibitem{panini}
Yinhuai Wang, Yujie Hu, and Jian Zhang.
\newblock Panini-net: Gan prior based degradation-aware feature interpolation
  for face restoration.
\newblock In {\em Proceedings of the AAAI Conference on Artificial Intelligence
  (AAAI)}, 2022.

\bibitem{wang2022zero}
Yinhuai Wang, Jiwen Yu, and Jian Zhang.
\newblock Zero-shot image restoration using denoising diffusion null-space
  model.
\newblock {\em International Conference on Learning Representations (ICLR)},
  2023.

\bibitem{whang2022deblurring}
Jay Whang, Mauricio Delbracio, Hossein Talebi, Chitwan Saharia, Alexandros~G
  Dimakis, and Peyman Milanfar.
\newblock Deblurring via stochastic refinement.
\newblock In {\em Proceedings of the IEEE/CVF Conference on Computer Vision and
  Pattern Recognition (CVPR)}, 2022.

\bibitem{zhang2021designing}
Kai Zhang, Jingyun Liang, Luc Van~Gool, and Radu Timofte.
\newblock Designing a practical degradation model for deep blind image
  super-resolution.
\newblock In {\em Proceedings of the IEEE/CVF International Conference on
  Computer Vision (ICCV)}, 2021.

\end{thebibliography}
}

\appendix

\section{DDNM for Noisy Image Restoration}
\label{app: ddnm noisy}
For noisy inverse problem in the form $\mathbf{y}=\mathbf{A}\mathbf{x}+\mathbf{n}$, $\mathbf{n}\sim\mathcal{N}(\mathbf{0},\sigma_{\mathbf{y}}^2\mathbf{I})$, DDNM uses the denoiser $\mathcal{Z}_{\boldsymbol{\theta}}$ to eliminate the external noise $\mathbf{n}$. To this end, DDNM involves two extra coefficients $\textcolor{blue}{\mathbf{\Sigma}_{t}}$ and $\textcolor{blue}{\boldsymbol{\Phi}_t}$, and turns Eq.~\ref{eq:ddnm rnd} and Eq.~\ref{eq:ddnm reverse} into
\begin{equation}
    \mathbf{\hat{x}}_{0|t} = \mathbf{x}_{0|t} + \textcolor{blue}{\mathbf{\Sigma}_{t}}\mathbf{A}^{\dagger}(\mathbf{y} - \mathbf{A}\mathbf{x}_{0|t}),
    \label{eq:ndm+ 1}
\end{equation}
\begin{equation}
\mathbf{x}_{t-1}=a_{t-1}\hat{\mathbf{x}}_{0|t}+\sigma_{t-1}(\textcolor{blue}{\boldsymbol{\Phi}_t}\boldsymbol{\epsilon}+\sqrt{1-\eta_t^2}\boldsymbol{\epsilon}_t),\quad \boldsymbol{\epsilon}\sim \mathcal{N}(0,\mathbf{I})
    \label{eq:ndm+ 2}
\end{equation}
The total noise distribution in $\mathbf{x}_{t-1}$ should be $\mathcal{N}(0,\sigma_{t-1}^2\mathbf{I})$ so that it can be removed by the denoiser $\mathcal{Z}_{\boldsymbol{\theta}}$:
\begin{equation}
a_{t-1}\textcolor{blue}{\mathbf{\Sigma}_{t}}\mathbf{A}^{\dagger}\mathbf{n}+\sigma_{t-1}(\textcolor{blue}{\boldsymbol{\Phi}_t}\boldsymbol{\epsilon}+\sqrt{1-\eta_t^2}\boldsymbol{\epsilon}_t)\sim\mathcal{N}(0,\sigma_{t-1}^2\mathbf{I})
    \label{eq:ndm+ 3}
\end{equation}
\begin{equation}
a_{t-1}\textcolor{blue}{\mathbf{\Sigma}_{t}}\mathbf{A}^{\dagger}\mathbf{n}+\sigma_{t-1}\textcolor{blue}{\boldsymbol{\Phi}_t}\boldsymbol{\epsilon}\sim\mathcal{N}(0,\sigma_{t-1}^2\eta_t^2\mathbf{I})
    \label{eq:ndm+ 4}
\end{equation}
Considering the variance equivalence:
\begin{equation}
a_{t-1}^2\sigma_{\mathbf{y}}^2\textcolor{blue}{\mathbf{\Sigma}_{t}}\mathbf{A}^{\dagger}(\textcolor{blue}{\mathbf{\Sigma}_{t}}\mathbf{A}^{\dagger})^{\top}+\sigma_{t-1}^2\textcolor{blue}{\boldsymbol{\Phi}_t}\textcolor{blue}{\boldsymbol{\Phi}_{t}^{\top}}=\sigma_{t-1}^2\eta_t^2\mathbf{I}
    \label{eq:ndm+ 5}
\end{equation}
As shown in Eq.~\ref{eq:ndm+ 5}, the coefficients $\textcolor{blue}{\mathbf{\Sigma}_{t}}$ and $\textcolor{blue}{\boldsymbol{\Phi}_t}$ are highly linearly coupled and are difficult to solve. So we need to use SVD to transform them into orthogonal space. The SVD of $\mathbf{A}$ and $\mathbf{A}^{\dagger}$ is:
\begin{equation}
\mathbf{A}=\mathbf{U\Sigma V}^{\top},\quad\mathbf{A}^{\dagger}=\mathbf{V\Sigma^{\dagger}U}^{\top}
    \label{eq:ndm+ 6}
\end{equation}

At the same time, we construct a special SVD for $\textcolor{blue}{\mathbf{\Sigma}_{t}}$ and $\textcolor{blue}{\mathbf{\Phi}_{t}}$ to further simplify Eq.~\ref{eq:ndm+ 5}.
\begin{equation}
\textcolor{blue}{\mathbf{\Sigma}_{t}}=\mathbf{V}\textcolor{blue}{\boldsymbol{\Lambda}_t}\mathbf{V}^{\top},\textcolor{blue}{\mathbf{\Phi}_{t}}=\mathbf{V}\textcolor{blue}{\boldsymbol{\Gamma}_t}\mathbf{V}^{\top}
    \label{eq:ndm+ 7}
\end{equation}
Then Eq.~\ref{eq:ndm+ 5} becomes
\begin{equation}
a_{t-1}^2\sigma_{\mathbf{y}}^2\mathbf{V}\textcolor{blue}{\boldsymbol{\Lambda}_{t}}\mathbf{\Sigma}^{\dagger}(\mathbf{\Sigma}^{\dagger})^{\top}\textcolor{blue}{\boldsymbol{\Lambda}_{t}}\mathbf{V}^{\top}+\sigma_{t-1}^2\mathbf{V}\textcolor{blue}{\boldsymbol{\Gamma}_{t}}^{2}\mathbf{V}^{\top}=\sigma_{t-1}^2\eta_t^2\mathbf{I}
    \label{eq:ndm+ 8}
\end{equation}
\begin{equation}
\mathbf{V}(a_{t-1}^2\sigma_{\mathbf{y}}^2\textcolor{blue}{\boldsymbol{\Lambda}_{t}}\mathbf{\Sigma}^{\dagger}(\mathbf{\Sigma}^{\dagger})^{\top}\textcolor{blue}{\boldsymbol{\Lambda}_{t}}+\sigma_{t-1}^2\textcolor{blue}{\boldsymbol{\Gamma}_{t}}^{2})\mathbf{V}^{\top}=\mathbf{V}\sigma_{t-1}^2\eta_t^2\mathbf{I}\mathbf{V}^{\top}
    \label{eq:ndm+ 8.1}
\end{equation}
\begin{equation}
a_{t-1}^2\sigma_{\mathbf{y}}^2\textcolor{blue}{\boldsymbol{\Lambda}_{t}}\mathbf{\Sigma}^{\dagger}(\mathbf{\Sigma}^{\dagger})^{\top}\textcolor{blue}{\boldsymbol{\Lambda}_{t}}+\sigma_{t-1}^2\textcolor{blue}{\boldsymbol{\Gamma}_{t}}^{2}=\sigma_{t-1}^2\eta_t^2\mathbf{I}
    \label{eq:ndm+ 9}
\end{equation}
The below matrices in Eq.~\ref{eq:ndm+ 9} are diagonal matrices:
\begin{equation}
\textcolor{blue}{\mathbf{\Lambda}_t}=diag\{\textcolor{blue}{\lambda_{t1}, \lambda_{t2}, \cdots, \lambda_{tD}}\}
    \label{eq:ndm+ 10}
\end{equation}
\begin{equation}
\textcolor{blue}{\mathbf{\Gamma}_t}=diag\{\textcolor{blue}{\gamma_{t1}, \gamma_{t2}, \cdots, \gamma_{tD}}\}
    \label{eq:ndm+ 11}
\end{equation}
\begin{equation}
\mathbf{\Sigma}^{\dagger}(\mathbf{\Sigma}^{\dagger})^{\top}=diag\{s_{1}^2, s_{2}^2, \cdots, s_{D}^2\}
    \label{eq:ndm+ 12}
\end{equation}
So Eq.~\ref{eq:ndm+ 9} is actually the equation on its diagonal elements:
\begin{equation}
a_{t-1}^2\sigma_{\mathbf{y}}^2\textcolor{blue}{\lambda_{ti}}^2 s_{i}^2+\sigma_{t-1}^2\textcolor{blue}{\gamma_{ti}}^2=\sigma_{t-1}^2\eta_t^2
    \label{eq:ndm+ 13}
\end{equation}
To make sure Eq.~\ref{eq:ndm+ 13} holds, we set
\begin{align}
    \textcolor{blue}{\gamma_{ti}}=\sqrt{\frac{\sigma_{t-1}^2\eta_t^2 - a_{t-1}^2\sigma_{\mathbf{y}}^2\textcolor{blue}{\lambda_{ti}}^2 s_{i}^2}{\sigma_{t-1}^2}}
    \label{eq:ndm+ 14}
\end{align}
To preserve the range-space information, we need $\mathbf{\Sigma}_t$ as close to $\mathbf{I}$ as possible. So we set
\begin{equation}
    \textcolor{blue}{\lambda_{ti}}=\begin{cases}
        1, & \sigma_{t-1}\eta_t \ge a_{t-1}\sigma_{\mathbf{y}}s_{i}, \\
        \frac{\sigma_{t-1}\eta_t}{a_{t-1}\sigma_{\mathbf{y}} s_{i}}, & \sigma_{t-1}\eta_t < a_{t-1}\sigma_{\mathbf{y}}s_{i}
    \end{cases},
    \label{eq:ndm+ 15}
\end{equation}
In this way, we calculate the coefficients $\textcolor{blue}{\mathbf{\Sigma}_{t}}$ and $\textcolor{blue}{\boldsymbol{\Phi}_t}$, by which DDNM can well solve noisy inverse problems. 

Note that in Eq.~\ref{eq:ndm+ 2}, the noise part can be also written as $\sigma_{t-1}\textcolor{blue}{\boldsymbol{\Phi}_t}(\boldsymbol{\epsilon}+\sqrt{1-\eta_t^2}\boldsymbol{\epsilon}_t)$ or $\sigma_{t-1}(\boldsymbol{\epsilon}+\textcolor{blue}{\boldsymbol{\Phi}_t}\sqrt{1-\eta_t^2}\boldsymbol{\epsilon}_t)$, if so, the calculation of $\textcolor{blue}{\mathbf{\Sigma}_{t}}$ and $\textcolor{blue}{\boldsymbol{\Phi}_t}$ will be different. 

\end{document}